\documentclass[conference, compsocconf]{IEEEtran}
\usepackage[cmex10]{amsmath}
\usepackage{url}
\usepackage{tikz-qtree}
\usepackage{graphicx}
\usepackage{caption}
\usepackage{fixltx2e}
\usepackage{float}
\usepackage{mathtools}

\DeclareMathOperator*{\score}{score}
\DeclareMathOperator*{\cnt}{count}
\DeclareMathOperator*{\fpr}{FPR}
\DeclareMathOperator*{\fnr}{FNR}

\begin{document}

\title{Content-based Approach \\for Vietnamese Spam SMS Filtering}

\author{\IEEEauthorblockN{Thai-Hoang Pham}
\IEEEauthorblockA{FPT Technology Research Institute\\
FPT University, Hanoi, Vietnam\\
Email: \textit{hoangpt@fpt.edu.vn}}
\and
\IEEEauthorblockN{Phuong Le-Hong}
\IEEEauthorblockA{College of Science\\
Vietnam National University, Hanoi, Vietnam\\
Email: \textit{phuonglh@vnu.edu.vn}}
}

\maketitle

\begin{abstract}
  Short Message Service (SMS) spam is a serious problem in Vietnam
  because of the availability of very cheap pre-paid SMS
  packages. There are some systems to detect and filter spam messages
  for English, most of which use machine learning techniques to
  analyze the content of messages and classify them. For Vietnamese,
  there is some research on spam email filtering but none focused on
  SMS. In this work, we propose the first system for filtering
  Vietnamese spam SMS. We first propose an appropriate preprocessing
  method since existing tools for Vietnamese preprocessing cannot
  give good accuracy on our dataset. We then experiment with vector
  representations and classifiers to find the best model for this
  problem. Our system achieves an accuracy of 94\% when labelling spam
  messages while the misclassification rate of legitimate messages is
  relatively small, about only 0.4\%. This is an encouraging result
  compared to that of English and can be served as a strong baseline
  for future development of Vietnamese SMS spam prevention systems.
\end{abstract}

\IEEEpeerreviewmaketitle

\section{Introduction}
In recent years, the explosion of the mobile network in Vietnam
stimulated the use of Short Message Service (SMS). Spam through SMS
has increased because the cost to the spammer is low and the response
rate is higher than email. SMS spam is a serious problem in Vietnam
because of the availability of unlimited and very cheap pre-paid SMS
packages. 

According to a report in 2015 by BKAV\footnote{www.bkav.com.vn}, an
information security firm, there were 13.9 million spam messages sent
to mobile phone users every day in Vietnam, and one out of two people
received spam messages~\cite{Bkav:2015}. Meanwhile, there were at
least 120 million active mobile users in the market according to a
report of the Ministry of Information and Communication, which was
released in 2015~\cite{Mic:2015}.

Spam SMS causes many issues for mobile users. They may suffer
financial loss from these messages by reacting to them. Users may
accidentally call to premium rate numbers or register for expensive
services by replying to these messages.  Moreover, they can be exposed
to some risks by accessing harmful websites or downloading
malwares. Mobile network operators also suffer financially because they
may lose users or spend more on spam prevention.

There are many methods and systems for filtering spam messages for
English in mobile networks~\cite{Guzella:2009, Delany:2012}. Most of
them use machine learning approaches to analyze the
content of messages and additional information to detect spam
messages. One of the first systems was proposed by
Graham~\cite{Graham:2002}, using Naive Bayes classifier to get a good
result. This work led to the development and application of other
machine learning algorithms to spam filtering. The state-of-the-art
system for English can detect spam SMS with an accuracy of 97.5\%
with a good false positive of only 0.2\%~\cite{Almeida:2011}.  

In this paper, we present the first system for Vietnamese spam SMS
filtering with a good accuracy. We first propose a method for the
preprocessing step because existing tools for Vietnamese preprocessing
cannot give good accuracy on our dataset. We then experiment with
vector representations and different classifiers to find the best model for this
problem. Our system achieves an accuracy of about 94\% when labelling spam
messages while the misclassification rate of legitimate messages is
relatively small, about 0.4\%. This is an encouraging result,
applicable in real-world applications and serves as a good baseline
for further improvement of future systems.

The paper is structured as follows. Section~\ref{sec:background}
introduces briefly the task of filtering spam SMS.
Section~\ref{sec:methodology} describes the methodology of our
system. Section~\ref{sec:experiment} presents our data and the
evaluation results and discussion. Finally,
Section~\ref{sec:conclusion} concludes the paper and suggests some
directions for future work.

\section{Background}\label{sec:background}

\subsection{Spam filtering system}
A SMS is a short text message that contains only a limited number of characters. The
structure of this message is divided into the header and the body. The
header contains some information about the subject, sender, and
recipient while the body is the actual content of the message that we
see on mobile phones. 

\begin{figure}[h]
\center
\includegraphics[scale=0.4]{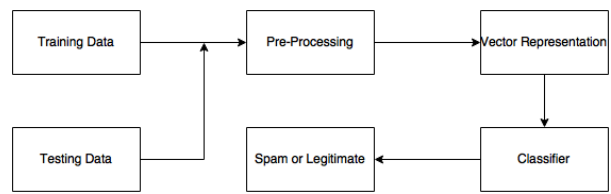}
\caption{Pipeline of a usual spam filter}\label{fig:1}
\end{figure}

Figure~\ref{fig:1} shows the pipeline of a usual spam filter. The
processing chain of a filter can be divided into three components. The
first is preprocessing, where documents are processed to extract data
for later use. The common preprocessing steps are tokenization,
lemmatization and removing stop words. After preprocessing the raw
text is transformed to vectors, each representing the content of one
message. The most popular representations are bag of words and term
frequency-inverse document frequency. The vectors are inputs to a
machine learning model. After be trained, the model is able to label
each new message as spam or not.

\subsection{Spam filtering system for Vietnamese}
To our knowledge, there is some research on spam filtering for
Vietnamese but none focused on SMS. Most of these systems are
developed for filtering spam email~\cite{Lung:2012,Dinh:2014}. The
best accuracy of those systems is about 92\% but the false positive
rate is not feasible in practical applications, about 6\%~\cite{Lung:2012}. Moreover, email
has many characteristics different from SMS, so applying these
systems directly to our task cannot give the best accuracy. For
these reasons, we propose a system that focuses on filtering
Vietnamese spam SMS.  

\section{Methodology}\label{sec:methodology}
SMS has some different characteristics compared to regular text such
as email, the task of labelling spam SMS is harder than for
email. Firstly the maximum length of an SMS message is only 160
characters, so the content-based approach for SMS is more difficult
because there is less information. Secondly, subscribers do not
usually use regular grammar to type a message. They use abbreviations,
phonetic contractions, wrong punctuations, emoticons, etc. It makes
the data sparse, and some preprocessing tools cannot be applied. Also,
linguistic characteristics of the Vietnamese language are different
from occidental languages. Thus, applying methods for English directly
or using existing systems for detecting Vietnamese spam email cannot
give the best performance. 

In the following subsections, we present our main method, including
preprocessing step, featurization and machine learning models. 

\subsection{Preprocessing}

\subsubsection{Entity Tagging}

The size of our corpus is relatively small and a lot of patterns in
messages are not words in the dictionary. Thus, it is necessary to
alleviate the sparseness of this data. We propose a new way to do this
by grouping some similar patterns together. We define six groups,
including  \textit{date, phone, link, currency, emoticon
  and number}. \textit{Date} collects characters that describe time
such as day, month, year, hour, etc\dots \textit{Phone} detects phone
numbers. \textit{Link} and \textit{currency} find and
replace characters that represent web links and money. Mobile users
use many \textit{emoticons} in messages and we match them to similar
emoticons of Facebook. After that, we replace all
number in the corpus by the token \textit{number}. 

\subsubsection{Tokenization}
Like many occidental languages, Vietnamese is based on the Latin 
alphabet. However in Vietnamese, the space symbol is not only used to
separate words but also to separate syllables within the same
word. The task of word segmentation plays an important role when
detecting spam message because it helps to extract features more
correctly. There are tools to solve this problem by detecting word
boundaries such as vnTokenizer~\cite{Le:2008},
JVnSegmenter~\cite{Nguyen:2006}. These tools work well for regular 
text but they are not suitable for our dataset because of SMS usually uses
an idiosyncratic language. Thus, we propose to use the phrase
collocation method for the segmentation task on our
dataset~\cite{Mikolov:2013}. The score of two syllables $w_{i}$,
$w_{j}$ that are parts of the same word is given by: 
\begin{equation}
\score(w_{i}, w_{j}) = \frac{\cnt(w_{i}w_{j} - \delta)}{\cnt(w_{i})*\cnt(w_{j})}
\end{equation}
where $\delta$ is used as a discounting coefficient to prevents too
many phrases consisting of very infrequent words to be formed.

For our system, we filter out all bigrams with total collected count
lower than 10. The advantage of this method is that we need no
pre-training data to train the tokenizer. It is based on only the
statistics of syllables in our corpus.

\subsection{Vector Representation}
The raw data type is text so we need to transform it to vector form to
use in machine learning models. There are two common methods for
converting from raw text to numeric vector - Bag of Words (BoW) and
Term frequency-Inverse document frequency (Tf-Idf). 

\subsubsection{Bag of words}
Given the vocabulary set $T = \{t_{1}, t_{2},..., t_{N}\}$, each
document $d$ in the corpus  is represented as a N-dimensional vector
$X = [x_{1}, x_{2},..., x_{N}]$, where $x_{i}$ is the
number of occurrences of $t_{i}$ in $d$. 

\subsubsection{Term frequency-Inverse document frequency}
Tf-Idf is a numerical statistic that is intended to reflect how
important a word is to a document in a collection or corpus. Now, each
$x_{i}$ in vector $X$ is calculated by an equation: 
\begin{equation}
x_{i} = n_{t_{i}, d} \log \left (\frac{|D|}{n_{t_{i}}} \right )
\end{equation}
where $D$ is a corpus, $n_{t_{i}, d}$ is the number of occurrences of
$t_{i}$ in $d$ and $n_{t_{i}}$ denotes the number of documents in
which $t_{i}$ occurs. 

\subsubsection{Feature Selection}
Because the represented vector is sparse, it may reduce the
performance of the classifier. To improve the accuracy of our system,
we need a method to select the most representative features such as
document frequency, information gain, term-frequency variance,
etc\dots In our setting, we use document frequency because this common
method has low computing cost.

\subsubsection{Length Feature}
In our experiment, we found that the length feature is useful because
the lengths of spam messages tend to be long. 

\subsection{Classifiers}

\subsubsection{Baseline Classifier}
Spam messages from mobile network operators are easy to detect because
they have special tags in their contents. Thus, we collect these tags
and label messages that have them as spam messages. Specifically,
these tags are \textit{[QC*], (QC*), [TB*], (TB*)} where * is a string
or empty. We use this method as a baseline system for the evaluation
task. 

\subsubsection{Machine Learning Models}
We trained some common machine learning models on our dataset and used
them to detect spam messages. These models include Naive Bayes (NB),
support vector machine (SVM), logistic regression (LR), decision tree
(DT) and k-nearest neighbours (kNN).

\section{Experiment}\label{sec:experiment}
\subsection{Dataset}
We conduct experiments on a dataset containing 6599 messages manually
annotated with labels. Our dataset is a collection of messages from
two biggest mobile network operators in Vietnam - Viettel and
Vinaphone. There are 5557 legitimate messages and 1042 spam messages
in this dataset. Its dictionary has $7,334$ unique words. Spam message
could be sent from either mobile network operators or other
sources. The former are easy to detect because they have special tags
in their contents. In our dataset about 70\% of spam messages are from
operators. The size of this corpus is comparable to some SMS datasets
for English~\cite{Delany:2012}. This dataset is made available for
research purpose\footnote{github.com/pth1993}.
 
\subsection{Results and Discussions}
\subsubsection{Evaluation Method}
We use 5-fold cross-validation to evaluate our system. The final
accuracy score is the average scores of the five runs.

We use statistical measures of the performance of a binary
classification test in decision theory as measures for our system. The
reason for using this method is the asymmetry in the misclassification
costs. The error that the system incorrectly classifies spam message
as legitimate is a minor problem while the error of labelling
legitimate message as spam can be unacceptable. Thus, there is a
trade-off between two types of errors. The sound system must satisfy
both of two constraint - the good performance for spam filtering and
the misclassification of the legitimate message is minimum. 

In decision theory, two classes are positive (spam) and negative
(legitimate). We measure two types of error by false positive rate and
false negative rate. The numbers of spam and legitimate message in our
dataset are $n_{S}$ and $n_{L}$, the numbers of incorrect messages are
$n_{S,L}$ and $n_{L,S}$, respectively. From these, the false positive
and negative rates are given by the following formulas:
\begin{equation}
\begin{split}
\fpr = \frac{n_{L,S}}{n_{L}}\\
\fnr = \frac{n_{S,L}}{n_{S}}
\end{split}
\end{equation}
\subsubsection{Baseline System}
Our baseline system uses a simple rule to detect spam messages by catching special tags in their contents. Table~\ref{tab:1} presents the accuracy of this system.
\begin{table}[h]
\center
\caption{Accuracy of baseline system}\label{tab:1}
\begin{tabular}{|c|c|}
\hline 
True Positive Rate & 69.76\% \\ 
\hline 
True Negative Rate & 100.00\% \\ 
\hline 
False Positive Rate & 0.00\% \\ 
\hline 
False Negative Rate & 30.24\% \\ 
\hline 
\end{tabular}
\end{table}

We see that this system can detect only about 70\% of the number of
spam messages because there are lots of spam messages that have no
special tags in their contents.  
\subsubsection{Preprocessing}
In the second experiment, we compare the performance of the system
with or without preprocessing. The classifier we use is Support Vector
Machine, and the representation is BoW. 

Without preprocessing the vocabulary has 7334 words while using
tokenization and entity tagging, the vocabulary size is reduced to
5998 words. It makes our corpus less sparse and more accurate because
its size is smaller and words in Vietnamese is combined from one or
more syllables. Table~\ref{tab:2} shows the improvement when using the
preprocessing technique. 
\begin{table}[h]
\center
\caption{Evaluation of preprocessing}\label{tab:2}
\begin{tabular}{|c|c|c|}
\hline
 & Without preprocessing & With preprocessing \\
\hline 
True Positive Rate & 91.79\% & 93.40\% \\ 
\hline 
True Negative Rate & 99.69\% & 99.60\% \\ 
\hline 
False Positive Rate & 0.31\% & 0.40\% \\ 
\hline 
False Negative Rate & 8.21\% & 6.60\% \\ 
\hline 
\end{tabular}
\end{table}

We see that with preprocessing the accuracy of labelling spam message
increases about 1.6\%.

\subsubsection{Vector Representations}
In the third experiment, we evaluate the performance of SVM with two common methods for
representing text -- BoW and Tf-Idf. The result is
shown in Table~\ref{tab:3}. 
\begin{table}[h]
\center
\caption{Evaluation of vector representations}\label{tab:3}
\begin{tabular}{|c|c|c|}
\hline
 & BoW & Tf-Idf \\
\hline 
True Positive Rate & 93.40\% & 84.66\% \\ 
\hline 
True Negative Rate & 99.60\% & 99.58\% \\ 
\hline 
False Positive Rate & 0.40\% & 0.42\% \\ 
\hline 
False Negative Rate & 6.60\% & 15.34\% \\ 
\hline 
\end{tabular}
\end{table}

From the information in above table, we can conclude that BoW is very
more useful than Tf-Idf for this problem. Specifically, the accuracy
when using BoW is better than using Tf-Idf about 15\%. 
\subsubsection{Classifiers}
In the fourth experiment, we compare some machine learning models to
find the best models for this task. Classifiers that we compare are
support vector machine (SVM), naive bayes (NB), decision tree (DT),
logistic regression (LR) and k-nearest neighbours (kNN) and the vector
representation in this experiment is BoW. Table~\ref{tab:4} shows the
accuracy of each model.
\begin{table}[h]
\center
\caption{Evaluation of classifiers}\label{tab:4}
\begin{tabular}{|c|c|c|c|c|c|}
\hline
 & SVM & NB & LR & DT & kNN \\
\hline 
True Positive Rate & 93.40\% & 95.15\% & 92.99\% & 92.18\% & 78.13\% \\ 
\hline 
True Negative Rate & 99.60\% & 96.25\% & 99.64\% & 99.12\% & 99.64\% \\ 
\hline 
False Positive Rate & 0.40\% & 3.75\% & 0.36\% & 0.88\% & 0.36\% \\ 
\hline 
False Negative Rate & 6.60\% & 4.82\% & 7.01\% & 7.82\% & 21.87\% \\ 
\hline 
\end{tabular}
\end{table}

Naive Bayes gives the best result for labelling spam message but its
false positive rate is very high, about 3.75\%. It means that for each
100 legitimate messages, there are 3 messages that are labelled as
spam. This rate is not acceptable in reality. In the rest, SVM gets
the best result. Thus, we use SVM as a classifier for our system. 

\subsubsection{Length feature and feature selection}
We find that the length feature is useful for our system because the
lengths of spam messages tend to be long. Document frequency is a good
method to find the most useful features for classifying and make our
system faster. Table~\ref{tab:5} presents the accuracy of our system
when adding length feature and set document frequency as 3. 
\begin{table}[h]
\center
\caption{Accuracy of our system}\label{tab:5}
\begin{tabular}{|c|c|}
\hline 
True Positive Rate & 93.91\% \\ 
\hline 
True Negative Rate & 99.55\% \\ 
\hline 
False Positive Rate & 0.45\% \\ 
\hline 
False Negative Rate & 6.09\% \\ 
\hline 
\end{tabular}
\end{table}

\section{Conclusion and Outlook}\label{sec:conclusion}
In this paper, we have presented the first system for Vietnamese spam
SMS filtering. Our system achieves a good accuracy of about 94\% of
detecting spam messages. This result is comparable to the accuracy of
this task for English and outperforms performances of some systems for
labelling Vietnamese spam email although classifying SMS is more
complicated.  

In the future, on the one hand, we plan to improve our system by enlarging
our corpus so as to provide more data for the system. On the other
hand, we would like to use some deep learning models such as
convolution neural network and long-short term memory for detect spam
messages because these models have been shown to give the best
performance for many text classification
problems~\cite{Johnson:2016,Zhang:2015}.

\section*{Acknowledgment}
The authors would like to thank the
CyRadar\footnote{cyradar.com} team for providing us the dataset for
use in the experiments. The second author is partly funded by the
Vietnam National University, Hanoi (VNU) under project number QG.15.04.

\bibliographystyle{IEEEtran}
\bibliography{IEEEabrv,mybib}

\begin{thebibliography}{10}
\providecommand{\url}[1]{#1}
\csname url@samestyle\endcsname
\providecommand{\newblock}{\relax}
\providecommand{\bibinfo}[2]{#2}
\providecommand{\BIBentrySTDinterwordspacing}{\spaceskip=0pt\relax}
\providecommand{\BIBentryALTinterwordstretchfactor}{4}
\providecommand{\BIBentryALTinterwordspacing}{\spaceskip=\fontdimen2\font plus
\BIBentryALTinterwordstretchfactor\fontdimen3\font minus
  \fontdimen4\font\relax}
\providecommand{\BIBforeignlanguage}[2]{{%
\expandafter\ifx\csname l@#1\endcsname\relax
\typeout{** WARNING: IEEEtran.bst: No hyphenation pattern has been}%
\typeout{** loaded for the language `#1'. Using the pattern for}%
\typeout{** the default language instead.}%
\else
\language=\csname l@#1\endcsname
\fi
#2}}
\providecommand{\BIBdecl}{\relax}
\BIBdecl

\bibitem{Bkav:2015}
\BIBentryALTinterwordspacing
B.~Corporation. (2015) The review of network security situation in 2015.
  [Online]. Available:
  \url{http://www.bkav.com.vn/hoi-dap/-/chi_tiet/383980/tong-ket-an-ninh-mang-nam-2015-va-du-bao-xu-huong-2016}
\BIBentrySTDinterwordspacing

\bibitem{Mic:2015}
\BIBentryALTinterwordspacing
M.~of~Information and Communication. (2015) The review of working in 2015.
  [Online]. Available:
  \url{http://mic.gov.vn/solieubaocao/Pages/TinTuc/116098/Tinh-hinh-phat-trien-linh-vuc-vien-thong--internet-nam-2015.html}
\BIBentrySTDinterwordspacing

\bibitem{Guzella:2009}
T.~S. Guzella and W.~M. Caminhas, ``A review of machine learning approaches to
  spam filtering,'' \emph{Expert Systems with Applications}, vol.~36, no.~7,
  pp. 10\,206--10\,222, 2009.

\bibitem{Delany:2012}
S.~J. Delany, M.~Buckley, and D.~Greene, ``{SMS} spam filtering: methods and
  data,'' \emph{Expert Systems with Applications}, vol.~39, no.~10, pp.
  9899--9908, 2012.

\bibitem{Graham:2002}
\BIBentryALTinterwordspacing
P.~Graham. (2002) A plan for spam. [Online]. Available:
  \url{http://www.paulgraham.com/spam.html}
\BIBentrySTDinterwordspacing

\bibitem{Almeida:2011}
J.~M. G.~H. Tiago A Almeida~and and A.~Yamakami, ``Contributions to the study
  of {SMS} spam filtering: new collection and results,'' in \emph{Proceedings
  of the 11th ACM symposium on Document engineering}, California, United
  States, 2011, pp. 259--262.

\bibitem{Lung:2012}
V.~D. Lung and T.~N. Vu, ``Bayesian spam filtering for {Vietnamese} emails,''
  in \emph{Proceedings of the International Conference on Computer \&
  Information Science}, Kuala Lumpur, Malaysia, 2012, pp. 190--193.

\bibitem{Dinh:2014}
Q.~D. Dinh, Q.~A. Tran, and F.~Jiang, ``Automated generation of ham rules for
  {Vietnamese} spam filtering,'' in \emph{Proceedings of the 7th IEEE Symposium
  on Computational Intelligence for Security and Defense Applications}, Hanoi,
  Vietnam, 2014, pp. 1--5.

\bibitem{Le:2008}
P.~Le-Hong, T.~M.~H. Nguyen, A.~Roussanaly, and T.~V. Ho, ``A hybrid approach
  to word segmentation of {V}ietnamese texts,'' in \emph{Language and Automata
  Theory and Applications}, ser. Lecture Notes in Computer Science.\hskip 1em
  plus 0.5em minus 0.4em\relax Springer Berlin Heidelberg, 2008, vol. 5196, pp.
  240--249.

\bibitem{Nguyen:2006}
N.~Cam-Tu, N.~Trung-Kien, P.~Xuan-Hieu, N.~Le-Minh, and H.~Quang-Thuy,
  ``Vietnamese word segmentation with {CRFs} and {SVMs}: An investigation,'' in
  \emph{Proceedings of the 20th Pacific Asia Conference on Language,
  Information and Computation}, Wuhan, China, 2006.

\bibitem{Mikolov:2013}
T.~Mikolov, I.~Sutskever, K.~Chen, G.~S. Corrado, and J.~Dean, ``Distributed
  representations of words and phrases and their compositionality,'' in
  \emph{Proceedings of the 26th Advances in Neural Information Processing
  Systems}, Nevada, United States, 2013, pp. 3111--3119.

\bibitem{Johnson:2016}
R.~Johnson and T.~Zhang, ``Supervised and semi-supervised text categorization
  using lstm for region embeddings,'' in \emph{Proceedings of the 33rd
  International Conference on Machine Learning}, New York, United States, 2016,
  pp. 526--534.

\bibitem{Zhang:2015}
X.~Zhang, J.~Zhao, and Y.~LeCun, ``Character-level convolutional networks for
  text classification,'' in \emph{Proceedings of the 28th Advances in Neural
  Information Processing Systems}, 2015, pp. 649--657.

\end{thebibliography}

\end{document}